\documentclass[
]{ceurart}

\usepackage{listings}
\usepackage{threeparttable}
\usepackage{tabularx}
\usepackage{color}
\usepackage{booktabs}
\usepackage{verbatim}
\usepackage{amsopn}
\usepackage{amsmath}
\usepackage{multirow}
\usepackage{colortbl}
\usepackage{makecell} 

\usepackage{eucal}
\usepackage{amsfonts}
\usepackage{bbm}
\usepackage{bm}

\usepackage{xspace}

\makeatletter
\DeclareRobustCommand\onedot{\futurelet\@let@token\@onedot}
\def\@onedot{\ifx\@let@token.\else.\null\fi\xspace}

\def\ie{\emph{i.e}\onedot}

\def\etal{\emph{et al}\onedot}
\makeatother

\begin{document}

\copyrightyear{2024}
\copyrightclause{Copyright for this paper by its authors.
  Use permitted under Creative Commons License Attribution 4.0
  International (CC BY 4.0).}

\conference{The 2nd Workshop \& Challenge on Micro-gesture Analysis for Hidden Emotion Understanding, Aug 3--9, 2024, Jeju, South Korea}

\title{Prototype Learning for Micro-gesture Classification}


\author[1]{Guoliang Chen}[%
orcid=0009-0002-7984-3184,
email=guoliangchen.hfut@gmail.com,
]

\author[1]{Fei Wang}[%
orcid=0009-0004-1142-6434,
email=jiafei127@gmail.com,
]

\author[4]{Kun Li}[%
orcid=0000-0001-5083-2145,
email=kunli.hfut@gmail.com,
]
\cormark[1]

\author[4]{Zhiliang Wu}[%
orcid=0000-0002-6597-8048,
email=wu_zhiliang@zju.edu.cn,
]

\author[4]{Hehe Fan}[%
orcid=0000-0001-9572-2345,
email=hehefan@zju.edu.cn,
]

\author[4]{Yi Yang}[%
orcid=0000-0002-0512-880X,
email=yangyics@zju.edu.cn,
]

\author[1,2,3]{Meng Wang}[%
orcid=0000-0002-3094-7735,
email=eric.mengwang@gmail.com,
]

\author[1,2,3,5]{Dan Guo}[%
orcid=0000-0003-2594-254X,
email=guodan@hfut.edu.cn,
]
\cormark[1]

\address[1]{School of Computer Science and Information Engineering, School of Artificial Intelligence, Hefei University of Technology (HFUT)}
\address[2]{Key Laboratory of Knowledge Engineering with Big Data (HFUT), Ministry of Education}
\address[3]{Institute of Artificial Intelligence, Hefei Comprehensive National Science Center, China}
\address[4]{CCAI, Zhejiang University, China}
\address[5]{Anhui Zhonghuitong Technology Co., Ltd.}

\cortext[1]{Corresponding author.}

\begin{abstract}
In this paper, we briefly introduce the solution developed by our team, HFUT-VUT, for the track of Micro-gesture Classification in the MiGA challenge at IJCAI 2024. The task of micro-gesture classification task involves recognizing the category of a given video clip, which focuses on more fine-grained and subtle body movements compared to typical action recognition tasks. Given the inherent complexity of micro-gesture recognition, which includes large intra-class variability and minimal inter-class differences, we utilize two innovative modules, \ie, the cross-modal fusion module and prototypical refinement module, to improve the discriminative ability of MG features, thereby improving the classification accuracy. Our solution achieved significant success, ranking 1st in the track of Micro-gesture Classification. We surpassed the performance of last year's leading team by a substantial margin, improving Top-1 accuracy by 6.13\%. 
\end{abstract}

\begin{keywords}
Micro-gesture \sep
action classification \sep
multi-model action recognition \sep
video understanding
\end{keywords}

\maketitle

\section{Introduction}
Micro-gestures (MGs)~\cite{liu2021imigue,chen2023smg} are defined as a special category of body gestures that are indicative of humans’ emotional status. Examples include ``head scratching'', ``nose rubbing'' and ``hand rubbing'', which are not intended to be communicative but arise spontaneously from stress or discomfort. Unlike indicative gestures, which are intended to facilitate communication, micro-gestures have not been well studied, and existing works are focused primarily on macro body movements, neglecting these subtle gestures and their connection to hidden emotions. 

Compared to ordinary action or gesture recognition, MGs present more challenges. MGs encompass more refined and subtle bodily movements that occur spontaneously during real-life interactions. Additionally, micro-gestures suffer from small inter-class differences and large intra-class differences. Therefore, micro-gestures that have unclear action boundaries, large individual differences, high similarity, and are easily influenced by the environment tend to be misclassified, which we refer to as ambiguous samples.

In this challenge, we adopt PoseConv3D~\cite{duan2022revisiting} as the baseline model, and the main contributions of our method are summarized as follows:
\begin{itemize}
\item We proposed a multi-modality-based network for micro-gesture classification. Specifically, we incorporate the cross-modal fusion module and prototypical refinement module for action classification. 
\item For the micro-gesture classification challenge, our method achieves a Top-1 accuracy of 70.254 on the iMiGUE test set. 
The experimental results indicate that our method effectively captures subtle changes in micro-gestures. 
\end{itemize}

\section{Methodology}

\begin{figure}[t]
\centering
\includegraphics[width=1.0\linewidth]{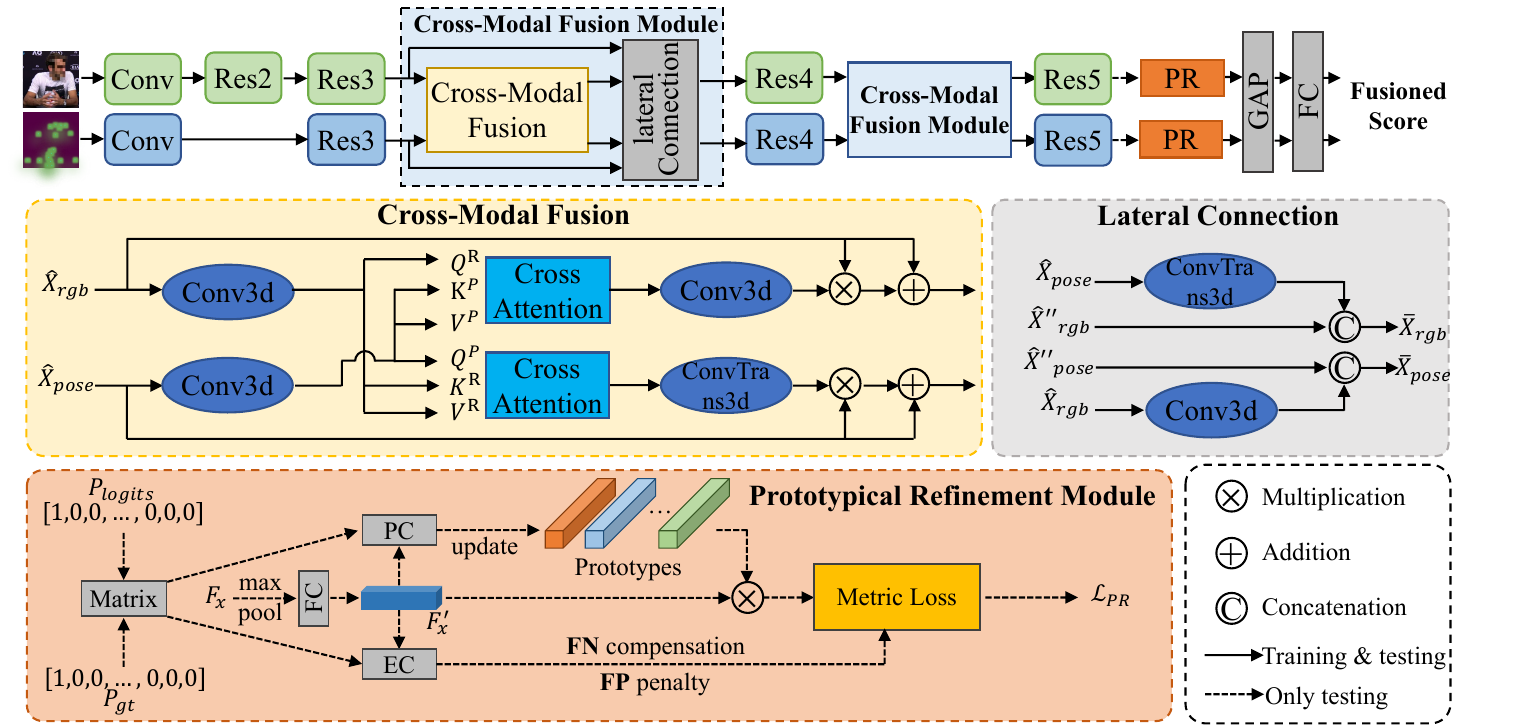}
\caption{Overview of the proposed method for micro-gesture classification. The network contains the RGB and Pose branches. The cross-attention fusion module is used to explore the correlations between RGB and Pose modalities. The prototypical refinement module defines prototype representations of each fine-grained micro-gesture category during training and forces the model to calibrate ambiguous samples among different micro-gesture categories. }
\label{fig:main}
\end{figure}

\subsection{Network Architecture}
The main structure of the proposed method is illustrated in Figure~\ref{fig:main}. 
We adopt the PoseConv3D~\cite{duan2022revisiting} network as the backbone, which enables more efficient learning of spatio-temporal features and greater robustness in noise handling. 
Concretely, the proposed method consists of a two-pathway 3D CNN-based backbone network, where the upper path is responsible for processing RGB data, and the lower path processes skeleton data. 

The inputs to the model are $X_{rgb}=(x_1, \ldots, x_{t_{rgb}})$ and $X_{pose}=(x_1, \ldots,x_{t_{pose}})$, where $t_{rgb}$ and $t_{pose}$ represent the number of frames, we set $t_{rgb}$ to 8 and $t_{pose}$ to 32. 
The backbone network comprises the Cross-Modal Fusion Module and Prototypical Refinement Module. 
Specifically, the Cross-Modal Fusion Module aims to explore the correlations between different modalities. 
The Prototypical Refinement Module defines prototype representations of each fine-grained micro-gesture category during training and forces the model to calibrate ambiguous samples among different micro-gesture categories. 
These feature vectors are subsequently mapped to the probability distribution of $K$ candidate categories through a fully connected (FC) layer with softmax activation. Finally, the probability distributions of the two modalities are fused in a 1:1 ratio to obtain the model's final prediction for the micro-gesture category.
 
\subsection{Cross-Modal Fusion Module}
Due to the subtle and complex nature of micro-gestures, different categories of micro-gestures may have different advantages and disadvantages in terms of expression in different modalities. 
To effectively leverage the characteristic features of different modalities within the backbone network, we propose the Cross-Modal Fusion module. 

Assuming the features of the module are $\hat{X}_{rgb} \in \mathbb{R}^{T_1 \times H_1 \times W_1 \times C_1}$ and $\hat{X}_{pose} \in \mathbb{R}^{T_2 \times H_2 \times W_2 \times C_2}$. 
Firstly, these features are mapped to the same hidden dimension through two 3D convolutional layers, and then a maximum pooling layer is used to compress the spatial features to (1,1). At this point, $\hat{X}_{rgb}, \hat{X}_{pose} \in (T_1, C')$, where $C'$ represents the hidden dimension. Subsequently, we utilize the cross-attention fusion mechanism to fuse the features learned from RGB and Skeleton modalities in a multi-level adaptive manner. 
We define the cross-attention fusion mechanism as:
\begin{equation}
\label{eq1}
\text{Attn}_m \in F_{attn}(Q^m, K^m, V^m), \text{where} \ m \in \{\text{RGB}, \text{Pose}\}. 
\end{equation}

The variables in Equation~\ref{eq1}  are defined as follows:
\begin{equation}
\begin{matrix}
Q^m = W^{q_m}\hat{X}_m^{'}, 
K^m = W^{k_m}\hat{X}_m^{'}, 
V^m = W^{v_m}\hat{X}_m^{'}. 
\end{matrix}
\end{equation}

In contrast to existing methodologies that implement attention~\cite{li2023transformer,tang2022gloss,li2023vigt} and cross-attention mechanisms~\cite{Wu_2023_CVPR,wu2024waveformer,10222097,10423816} along spatial dimensions, this paper introduces a novel approach that applies cross-attention based on channel dimensions. This technique enables the model to execute information fusion across different modalities, specifically at the channel level, and facilitates the learning of correlations between channel-dimensional information across these modalities. Additionally, it effectively minimizes redundant and noisy data within the channels. By focusing cross-attention on channel dimensions rather than the entire feature map, this method significantly reduces computational complexity, offering a more efficient solution for processing multimodal data.

Finally, the resulting attentional weights are multiplied with the inputs in the channel dimension and then summed: 
\begin{equation}
\left\{\begin{aligned}
\hat{X}_{rgb}^{''}&=(Conv3d(\hat{X}_{rgb}^{'})\cdot \hat{X}_{rgb})+\hat{X}_{rgb}, \\  
\hat{X}_{pose}^{''}&=(ConvTrans3d(\hat{X}_{pose}^{'})\cdot \hat{X}_{pose})+\hat{X}_{pose}.
\end{aligned}
\right.
\end{equation}

Finally, the fused $\hat{X}_{rgb}^{''}$ and $\hat{X}_{pose}^{''}$ are spliced through lateral connections:
\begin{equation}
\left\{\begin{aligned}
\bar X_{rgb}  &= [Conv3d(\hat{X}_{pose}); \hat{X}_{rgb}^{''}], \\  
\bar X_{pose} &= [ConvTrans3d(\hat{X}_{rgb}); \hat{X}_{pose}^{''}],
\end{aligned}
\right.
\end{equation}
where$[~;~]$ denotes the concatenation of two features.

\subsection{Prototypical Refinement Module}
Micro-gesture recognition suffers from the problem of large intra-class differences and small inter-class differences. 
In order to solve this problem, we utilize the prototypical refinement module inspired by~\cite{zhou2023learning}, which uses prototype-based contrastive learning, and is dedicated to discovering and calibrating the ambiguous samples between these categories.
As shown in Figure~\ref{fig:main}, we firstly utilize backbone network to extract video feature as $F_x \in \mathbb{R}^{N \times C\times H \times W}$, and preliminary categorical predictions are calculated as $P_{logits}\in \mathbb{R}^{N \times K}$, where $K$ represents the number of action categories for micro-gestures. The ground-truth is also obtained as $P_{gt}\in \mathbb{R}^{N \times K}$. 
The extracted feature $F_x$ is then fed into a maximum pooling layer and a fully connected layer, and then we can get $F^{'}_x\in\mathbb{R}^{N \times D}$, where $D$ is the hidden dimension.

\textbf{Ambiguous Samples Discovery.} 
Given an action label $k$, using $P_{logits}$ and $P_{gt}$, we can find the confidence samples and ambiguous samples in the training phase as follows. 
If a sample is predicted correctly, namely as a True Positive (TP), we consider it a confident sample to distinguish it from ambiguous samples. 
If a sample of action $k$ is misclassified as another category, it is called a False Negative (FN). If samples from other categories are misclassified as action $k$, it is called False Positive (FP). We find these ambiguous samples in a batch and compute their center representations as follows:
\begin{equation}
\mu_{FN}^k= \frac{1}{n_{FN}^k} \sum\limits_{j \in s_{FN}^k} F_j, \mu_{FP}^k=\frac{1}{n_{FP}^k} \sum\limits_{j \in s_{FP}^k} F_j.
\end{equation}

\textbf{Prototype Clustering.}
We created a prototype representation for each micro-gesture category and randomly initialized it before training. Take the action category $k$ as an example, the prototypes $\mathcal{P}_k$ serve as a stable estimate of the clustering center for action category $k$ and can be continuously optimized during the training process through confidence samples (namely true positive TP samples) of category $k$. With online training, the prototype $\mathcal{P}_k$ of action category $k$ can be updated by exponential moving average (EMA):
\begin{equation}
    \mathcal{P}_k=(1-\rho)\cdot  \frac{1}{n_{TP}^k} \sum\limits_{i \in s_{TP}^k}F_i+\rho \cdot \mathcal{P}_k^{pre},
\end{equation}
where $F_i$ is the feature of sample $i$, $\mathcal{P}_k^{pre}$ is the prototype before updating. $\rho$ is a momentum term and is empirically set to 0.9. 

\textbf{Error Calibration.} 
To calibrate the prediction of ambiguous samples, we use the confidence sample $i$ of action $k$ as an anchor and compute the two auxiliary terms in the feature space, which are $\phi_i$ and $\varphi_i$ for FN and FP samples, respectively, and are defined as follows:
\begin{equation}
    \phi_i=\left\{\begin{matrix} 
  1-dis(F_i,\mu_{FN}^k), if\ i \in s_{TP}^k \ and \  n_{FN}^k>0 \\  
  0, \ otherwize 
\end{matrix}\right. ,
\end{equation}
\begin{equation}
    \varphi_i=\left\{\begin{matrix} 
  1+dis(F_i,\mu_{FP}^k), if\ i \in s_{TP}^k \ and \  n_{FP}^k>0 \\  
  0, \ otherwize 
\end{matrix}\right. ,
\end{equation}
where $dis(,)$ denotes the cosine distance between two features.
$\phi_i$, as the compensation term for the set of FN samples, should be closer to the confidence samples in the feature space by minimizing $\phi_i$, motivating the model to correct the ambiguous samples as action $k$. $\varphi_i$, as the penalty term for the FP samples, should be further away from the confidence samples in the feature space by minimizing $\varphi_i$, preventing the model from identifying the ambiguous samples as action $k$.  
Finally, its prototypical refinement loss is defined as follows:
\begin{align}
\begin{aligned}
    \mathcal{L}_{PR}(i) & = - \log{ \frac{\mathcal{e}^{dis(F_i,P_k)/\tau -(1-p_{ik})\varphi_i}}{\mathcal{e}^{dis(F_i,P_k)/\tau-(1-p_{ik})\varphi_i}+ \sum\limits_{l \neq k}\mathcal{e}^{dis(F_i,P_l)/\tau} } },\\
&- \log{ \frac{\mathcal{e}^{dis(F_i,P_k)/\tau -(1-p_{ik})\phi_i}}{\mathcal{e}^{dis(F_i,P_k)/\tau-(1-p_{ik})\phi_i}+ \sum\limits_{l \neq k}\mathcal{e}^{dis(F_i,P_l)/\tau} } },
\end{aligned}
\end{align}
where $p_{ik}$ is the predicted probability score of sample $i$ for the micro-gesture category $k$. 

Finally, the total loss in the training process can be defined as follows:
\begin{equation}
\mathcal{L}=\mathcal{L}_{CE}+\alpha \cdot \mathcal{L}_{PR},
\end{equation}
where $\mathcal{L}_{CE}$ is Cross-Entropy loss and $\alpha$ is a hyper-parameter.

\section{Experiments}
\subsection{Datasets}
\textbf{iMiGUE~\cite{liu2021imigue} dataset.} This dataset comprises 32 micro-gestures, along with one non-micro-gesture class, collected from post-match press conference videos of tennis players. 
This challenge follows a cross-subject evaluation protocol, wherein the 72 subjects are divided into a training set consisting of 37 subjects and a testing set comprising 35 subjects. 
For the MG classification track, 12,893, 777, and 4,562 MG clips from iMiGUE are used for train, val, and test, respectively. 
In this challenge, it is allowed to use RGB and skeleton modal data. 

\subsection{Evaluation Metrics and Implementation Details}
For the micro-gesture classification challenge, we calculate the Top-1 Accuracy to assess the prediction results. 
MMAction2~\cite{2020mmaction2} toolbox has been used for the implementation of our approach. 
Since we use PoseConv3D~\cite{duan2022revisiting} as the baseline method, we first need to train the RGB and Pose branches individually and fuse the best weights obtained from these two modalities as the initial weights for the two-branch method. The frame sizes for RGB and poses are empirically set to 8 and 32, respectively. 
The Stochastic Gradient Descent (SGD) optimizer is employed with a momentum of 0.9 and a weight decay of 1e-4 in the training process. We set the batch size to 10 and the initial learning rate to 0.0075. The learning rate is reduced by a factor of 10 at the 8-th and 22-th epochs, and the model is trained with 30 epochs. 

\subsection{Experimental Results}
As shown in Table~\ref{tab:kaggle_results}, we first report top-3 results on the test set of the iMiGUE dataset. Our team achieves the best Top-1 Accuracy of 70.254. Compared to the team of ``ywww11'', our method outperforms it by 1.94\%. 
In addition, we also compare our approach with skeleton-based and RGB-based action recognition methods on the iMiGUE datasets. 

\begin{table}[t!]
\footnotesize
\caption{The top-3 results of micro-gesture classification on the iMiGUE test set. Data is provided by the Kaggle competition page\protect\footnotemark[1]. 
}
\begin{tabular}{c|cc}
\toprule
Rank & Team & Top-1 Accuracy (\%) \\ \hline
1 & HFUT-VUT (\textbf{Ours}) & \textbf{70.254}  \\ \hline
2 & NPU-MUCIS & 70.188 \\
3 & ywww11 & 68.917 \\
\bottomrule
\end{tabular}
\label{tab:kaggle_results}
\end{table}
\footnotetext[1]{The Kaggle competition page: \href{https://www.kaggle.com/competitions/2nd-miga-ijcai-challenge-track1/leaderboard}{https://www.kaggle.com/competitions/2nd-miga-ijcai-challenge-track1/leaderboard}}

\begin{table}[t!]
\caption{The results of micro-gesture classification on the test set of the iMiGUE dataset. $\boldsymbol{J}$ and $\boldsymbol{L}$ denote the joint and limb skeleton data, respectively.} 
\begin{tabular}{c|c|c}
\toprule
Method &Modality & Top-1 Accuracy (\%) \\ \hline
TSM~\cite{lin2019tsm} & RGB & 58.77 \\
VSwin-T ~\cite{liu2022video} & RGB & 59.97 \\ 
VSwin-S ~\cite{liu2022video} & RGB & 57.83 \\
VSwin-B ~\cite{liu2022video} & RGB & 61.73 \\ \hline
ST-GCN~\cite{yan2018spatial}   & Skeleton ($\boldsymbol{J}$) & 46.38 \\
ST-GCN++~\cite{duan2022pyskl}  & Skeleton ($\boldsymbol{J}$) & 49.56 \\
StrongAug~\cite{duan2022pyskl} & Skeleton ($\boldsymbol{J}$) & 53.13 \\
AAGCN~\cite{shi2020skeleton}   & Skeleton ($\boldsymbol{J}$) & 54.73 \\
CTR-GCN~\cite{chen2021channel} & Skeleton ($\boldsymbol{J}$) & 53.02 \\
DG-STGCN~\cite{duan2022dg}     & Skeleton ($\boldsymbol{J}$) & 49.56\\ 
PoseConv3D~\cite{duan2022revisiting} & Skeleton ($\boldsymbol{J}$) & 61.11 \\ 
\hline
Li~\etal (MiGA'23 1st)~\cite{li2023joint} & Skeleton ($\boldsymbol{J}$ + $\boldsymbol{L}$) & 64.12 \\
EHCT (MiGA'23 2nd)~\cite{huang2023micro} & Skeleton ($\boldsymbol{L}$) &63.02 \\ \hline
Ours (PoseConv3D) & Skeleton ($\boldsymbol{J}$) & 67.91 \\
Ours (Ensemble Model) & RGB \& Skeleton ($\boldsymbol{J}$ + $\boldsymbol{L}$) &\textbf{70.25}\\
\bottomrule
\end{tabular}
\label{tab:main}
\end{table}
As shown in Table~\ref{tab:main}, we explore the action recognition methods using different modalities on the test set of the iMiGUE dataset. 
For the RGB-based methods, compared to TSM~\cite{lin2019tsm}, Video Swin Transformer~\cite{liu2022video} is able to capture the local feature information better by local self-attention operation and achieves an excellent Top1-Accuracy result of 61.73. 
For the skeleton-based method, Li~\etal~\cite{li2023joint} incorporates a semantic embedding loss to improve action classification performance and achieves the Top-1 accuracy of 64.12. 
Huang~\etal~\cite{huang2023micro} propose an ensemble hypergraph-convolution Transformer equipped with an auxiliary classifier to mitigate the impact of imbalanced data. 
In contrast, our method utilizes a multi-modal fusion strategy based on RGB and skeleton modalities to bridge the information differences between modalities and capture the information of micro-gestures more comprehensively. 
Since there are very similar features between different micro-gesture categories, the prototypical refinement module is used to calibrate these ambiguous micro-gesture samples. 
As a result, our method achieves 67.91 on the Top-1 accuracy. 
To incorporate the advantages of RGB-based and skeleton-based methods, we perform a model ensemble on the results of the Video Swin Transformer~\cite{liu2022video} base version and Li~\etal~\cite{li2023joint} and our method.  
Finally, our method achieves 70.25 on the Top-1 accuracy, which is 6.13\% higher than the results of MiGA’23 1st place. 

\section{Conclusion}
In this paper, we present an innovative approach based on prototype learning for micro-gesture recognition in the MiGA challenge hosted at IJCAI 2024. Our approach adopts the PoseConv3D~\cite{duan2022revisiting} model as the baseline, incorporating the cross-modal fusion module and prototypical refinement module for action classification. 
The RGB modality contains rich texture information, and the pose modality provides geometric information about the human structure. Through the cross-modal fusion module, the model can establish a correlation between the two modalities and combine the information of both modalities. 
The prototypical refinement module can provide a stable and clear representation of each category by creating a prototype for each micro-gesture action category. This module effectively mitigates intra-category differences in micro-gestures and makes the distinction between categories clearer. 
Eventually, the proposed approach reached 70.254 on the test set of the iMiGUE dataset. 

In future work, we will address the issues in this challenge from other perspectives. One promising direction is the application of video motion magnification techniques~\cite{wang2024eulermormer,wang2024frequency} to magnify the subtle changes of micro-gestures for better recognition. 
Additionally, we plan to pre-train our model on the large-scale micro-action recognition dataset proposed by~\cite{guo2024benchmarking,li2024mmad} to investigate the impact of prior knowledge for micro-gesture recognition. 
We also plan to exploit the weakly supervised method~\cite{zhou2024advancing} or unsupervised method~\cite{wei2021deraincyclegan} to alleviate the difficulty of data annotation in micro-gesture recognition. 

\begin{acknowledgments}
This work was supported by the National Key R\&D Program of China (No. 2022YFB4500601), the National Natural Science Foundation of China (No. 62272144,72188101,62020106007 and U20A20183), the Fundamental Research Funds for the Central Universities (No. 226-2022-00051, JZ2024HGTG0309), and the Major Project of Anhui Province (No. 202203a05020011). 
\end{acknowledgments}

\bibliography{sample-ceur}


\end{document}